\documentclass[11pt,a4paper]{article}
\usepackage[hyperref]{acl2020}
\usepackage{times}
\usepackage{latexsym}

\usepackage{microtype}
\usepackage{multirow}
\usepackage{booktabs}
\usepackage{amsmath}
\usepackage{amsfonts,amssymb}
\usepackage{makecell}
\usepackage{subfigure}
\usepackage{graphicx}
\usepackage{enumitem}
\usepackage{amsfonts}
\usepackage{array}
\usepackage[ruled,linesnumbered]{algorithm2e}

\aclfinalcopy 

\title{Chat as Expected: Learning to Manipulate \\Black-box Neural Dialogue Models}

\author{Haochen Liu \\
  Michigan State University\\
  \texttt{liuhaoc1@msu.edu} \\
  \\
  \textbf{Tyler Derr} \\
  Michigan State University\\
  \texttt{derrtyle@msu.edu} \\
   \\\And
  \textbf{Zhiwei Wang} \\
  Michigan State University\\
  \texttt{wangzh65@msu.edu} \\
  \\
  \textbf{Jiliang Tang} \\
  Michigan State University\\
  \texttt{tangjili@msu.edu} \\}

\date{}

\begin{document}

\maketitle
\begin{abstract}
Recently, neural network based dialogue systems have become ubiquitous in our increasingly digitalized society. However, due to their inherent opaqueness, some recently raised concerns about using neural models are starting to be taken seriously. In fact, intentional or unintentional behaviors could lead to a dialogue system to generate inappropriate responses.
Thus, in this paper, we investigate whether we can learn to craft input sentences that result in a black-box neural dialogue model being manipulated into having its outputs contain target words or match target sentences. We propose a reinforcement learning based model that can generate such desired inputs automatically.
Extensive experiments on a popular well-trained state-of-the-art neural dialogue model show that our method can successfully seek out desired inputs that lead to the target outputs in a considerable portion of cases.
Consequently, our work reveals the potential of neural dialogue models to be manipulated, which inspires and opens the door towards developing strategies to defend them.
\end{abstract}

\section{Introduction}
In recent years, we have seen an astonishingly fast adoption of dialogue systems being utilized across many domains and we are interacting with them more and more in our everyday lives. From early such as chatbots ELIZA~\cite{weizenbaum1966eliza} and ALICE~\cite{wallace2001artificial} to later ones like Siri and XiaoIce~\cite{shum2018eliza}, the techniques have evolved from hand-crafted rule-based methods \cite{weizenbaum1966eliza,goddeau1996form}, retrieval-based methods \cite{lu2013deep,hu2014convolutional} to learning-based methods \cite{shang2015neural,vinyals2015neural,serban2016building}.
Recently, because of the advent of a series of deep learning models and the appearance of large-scale real dialogue corpora, end-to-end neural generative dialogue models emerge~\cite{chen2017survey}.
Due to the simple implementation and the strong generalization ability of neural dialogue models, they are one of the most popular techniques to build practical chatbot services \cite{DBLP:journals/corr/abs-1812-08989}.

However, with the wide application of neural dialogue models, the ethical challenges they bring are attracting more and more attention~\cite{henderson2017ethical,liu2019does,dinan2019queens}. More recently it has been demonstrated that neural networks suffer from some problems including vulnerability due to their black-box nature and our lack of truly understanding their inner processing~\cite{szegedy2013intriguing,DBLP:journals/corr/abs-1909-08072}. Thus, as we are integrating these systems into more critical and sensitive domains, it raises some serious concerns~\cite{yuan2019adversarial}, such as whether or not they can be manipulated to provide certain desired output~\cite{he2018detecting,liu2019say}.
More specifically, if such systems can be manipulated it could potentially cause a drastic shift to the current paradigm of how we interact with these systems.
For example, Tay, an AI chatbot developed by Microsoft, was shut down shortly after release due to its racism, sexist and obscene speech~\cite{wolf2017we}.
Online troublemakers found its vulnerability and tried a variety of inputs to induce it to output inappropriate (malicious or sensitive) responses \cite{price2016microsoft}.
To that end, in this work, we set out to study this fundamental question of whether we can learn to manipulate state-of-the-art black-box dialogue models to produce target outputs by crafting inputs automatically. It goes without saying that if indeed we can manipulate these systems and if we are currently integrating them into our daily lives at such a rapid pace, then this opens the door to a plethora of potential harmful attacks that could be performed and potentially result in an almost unimaginable set of possible negative outcomes.

Nevertheless, even having now realized how critical this question is to being answered, the path to discovering an answer has numerous challenges.
First, unlike many existing studies in other domains such as images~\cite{goodfellow2014explaining,chen2018attacking}, here the input search space is discrete and thus the traditional gradient-based optimization methodologies cannot be harnessed effectively~\cite{zhang2019survey,zhao2017generating}.
Furthermore, while seeking to discover if the current dialogue systems can be manipulated, we should not make the unreasonable assumption of having access to the full details/knowledge (i.e., model structure and parameters) of the system.
Currently, most developed methodologies have focused on the continuous input space domain and furthermore assumed access to the model.
Thus, since our problem is defined with a discrete input domain and our concern of whether these models can be manipulated is more realistic in the setting of a black-box dialogue model, then existing methods can not be applied 
and we require the development of novel frameworks to answer this indispensable fundamental question.


To address the above-mentioned challenges, in this paper, we regard the learning to craft input sentences as a sequential decision-making process. To this end, we propose the Target Dialogue Generation Policy Network (TDGPN), which serves as a reinforcement learning (RL) agent to iteratively generate tokens guided towards specific objectives. The proposed policy networks are optimized by the REINFORCE-style estimators~\cite{williams1992simple}, eliminating the needs for standard gradient back-propagation which is largely impaired by the discrete nature of the sentence generation process and the assumption of no access to the dialogue model parameters. Our main contributions are summarized as follows:
\begin{itemize}[leftmargin=0.5cm]
\item We identify the importance of exploring the potential of a black-box neural dialogue model to be manipulated towards a target output. 

\item We devise an RL-based framework TDGPN to effectively overcome the challenges associated with crafting inputs that enable black-box neural dialogue models to generate target responses.

\item Experiments on a well-trained black-box neural dialogue model verify the effectiveness of TDGPN that lead to target responses with a considerable success rate.
\end{itemize}

\section{The Proposed Framework}
\label{sec:model}
\begin{figure}[t]
\begin{center}
\includegraphics[scale=0.48]{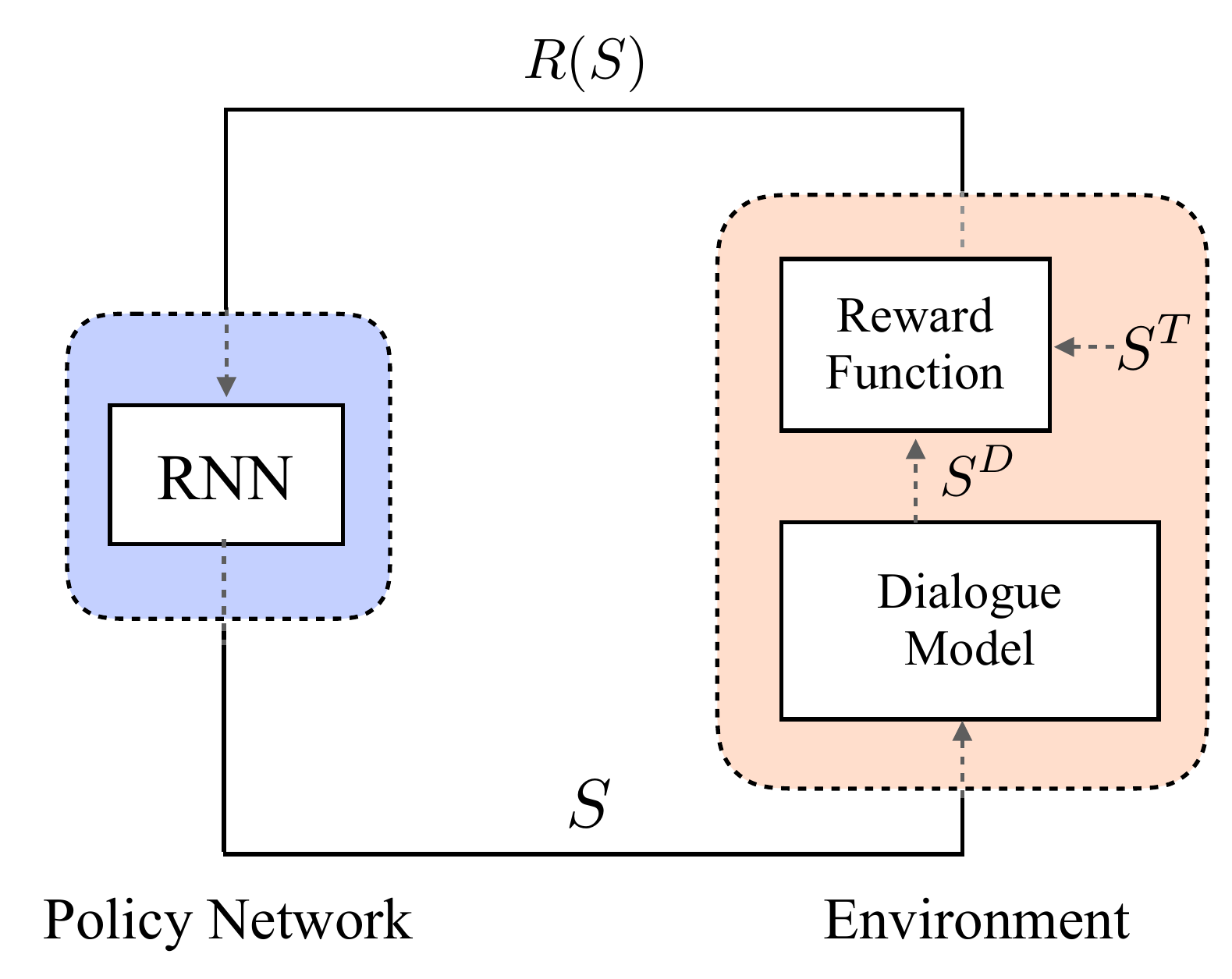}
\end{center}
\vspace{-2ex}
\caption{Diagram showing the overall reinforcement learning process for TDGPN.} 
\label{fig:diag}
\vskip -3ex
\end{figure}

\subsection{Problem Definition}
Before detailing the proposed framework, we first recap the problem of learning to manipulate neural dialogue models.
Suppose we have a neural dialogue model $D$, which is regarded as a black box and able to output a response sentence $S^D$ for a given input sentence. We seek for learning to craft an input sentence $S$ that will lead $D$ to output a response $S^D:=D(S)$ that satisfies specific requirements. We consider the following two manipulation tasks:

\textbf{Target word task.} Given a target word $w^T$, we aim to learn to craft an input sentence $S$ that will lead the dialogue model to output a response sentence $S^D$ that contains the target word, i.e., $w^T \in S^D$.

\textbf{Target response task.} Given a target response $S^T$, we aim to learn to craft an input sentence $S$ that will lead the dialogue model to output a response sentence $S^D$ that is semantically approximate to the target response, i.e., $S^D \approx S^T$.

We build a generative model $G_{\theta}$ parameterized by $\theta$ to learn to craft such input sentences.
Then the above problems are both essentially an optimization problem where we want to find optimum parameters of $G$ so that a) the probability of $w^T$ being generated in $S^D$, or b) the similarity between $D(S)$ and $S^T$ can be maximized. However, it is very challenging to solve this problem with the standard gradient-based back-propagation methods because of two reasons. First, $S$ consists of discrete tokens instead of continuous values. Thus, it is difficult to let the gradient back through G. Second, we do not have access to the dialogue model parameters, which makes it impossible to compute the gradient w.r.t $D$. Therefore, in this section, we formulate the problem as an RL problem and represent the generative model $G_{\theta}$ by a policy network $\pi_{\theta}$, which regards the token generation process as a decision making process with the purpose to obtain a maximum reward.

The overall learning diagram is shown in Figure~\ref{fig:diag}. There are two key components, i.e., policy network and environment which consists of a dialogue model and a reward function. The policy network will interact with the environment by generating input sentences $S$. Then, based on $S$ the environment will provide rewards to the policy network, which will in turn update its parameters towards obtaining maximum rewards. Next, we detail the interaction process.

\begin{figure}[t]
\begin{center}
\includegraphics[scale=0.3]{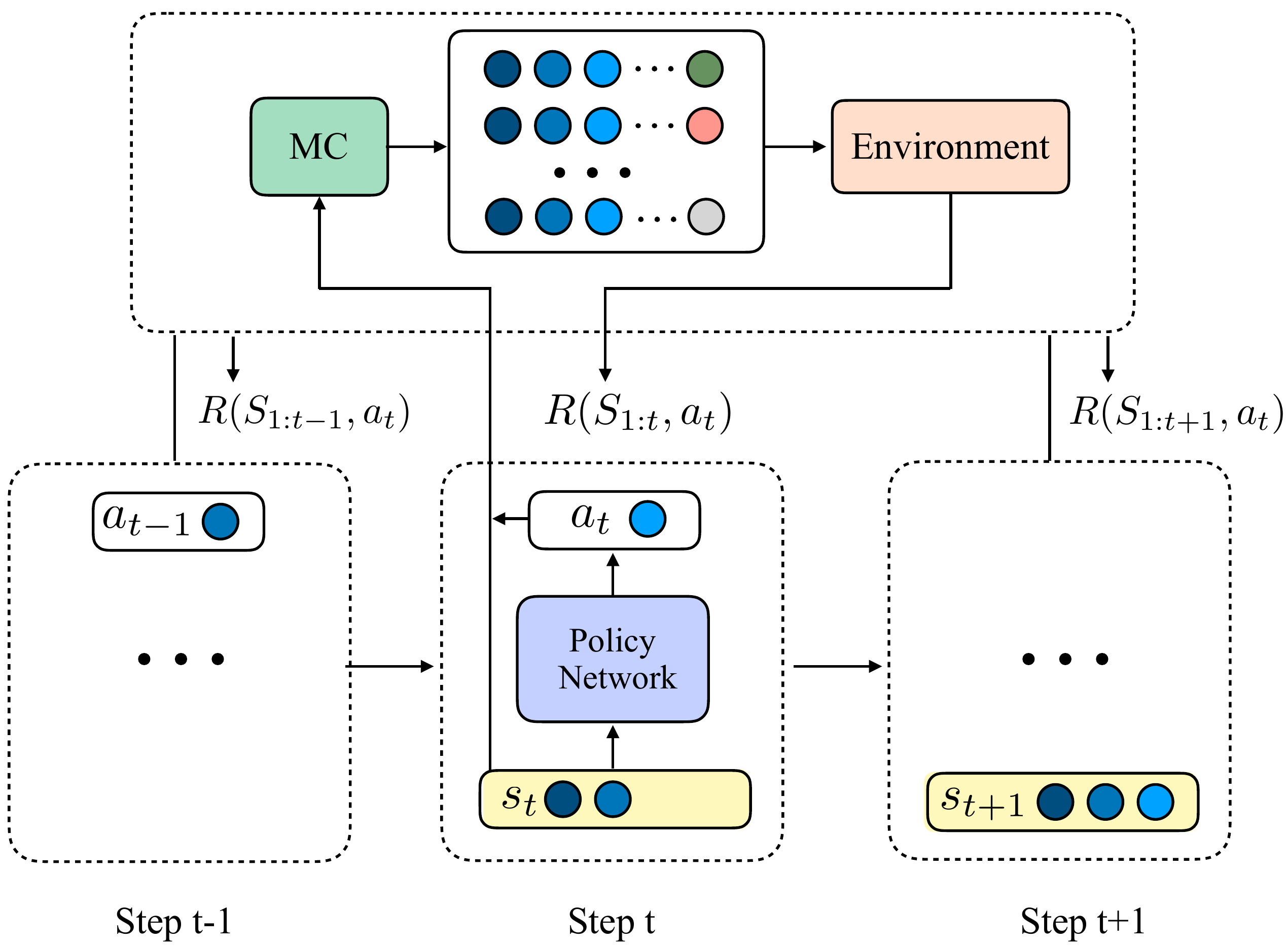}
\end{center}
\vspace{-1ex}
\caption{Diagram showing how to obtain the intermediate reward $R(S_{1:t},a_t)$ at step $s_t$ using the Monte Carlo (MC) method.} 
\label{fig:rewards}
\vskip -3ex
\end{figure}

\subsection{Sentence Generation Process}

In this subsection, we describe the sentence generation process which can be regarded as a decision making process. On a high level, at each step, a policy network observes the state,  outputs a token (i.e., makes an action) and receives a reward based on the dialogue model (i.e. environment). The generation process will be terminated when the policy network outputs an {\it end-of-sentence} token, after which the policy network will be updated according to the rewards. Next, we describe the state, actions, policy network, reward, and objective function. 
 
\subsubsection{State and Action}
In our framework, we denote the state at step $t$ of the sentence generation process as $s_{t}$ and let $s_t = S_{1:t}$, where $S_{1:t}$ is a sequence of already generated tokens at step $t$. More specifically,  $S_{1:t} = \{x_1, x_2, \cdots, x_t\}$, where $x_i \in \mathcal{V}$ and $\mathcal{V}$ is the vocabulary containing all the possible tokens. In addition, we assume that the state is fully observed. The start state is $s_0=\{x_0\}$ which consists of a special token $x_0$ that indicates the start of a sentence. Thus, the state transition is deterministic such that $P(s_t=S_{1:t}| s_{t-1}, a_t=x_t) = 1$ and $P(s_t=s\prime| s_{t-1}, a_t=x_t) = 0, \forall s\prime\neq S_{1:t}$, where $a_t$ is the action (i.e., token) at step $t$.
 
\subsubsection{Policy Network}
In this work, we represent the policy by a Long Short-Term Memory (LSTM) Network whose input is the state and output is selection probabilities $p$ over all the possible actions, which are the tokens in $\mathcal{V}$ including a special token indicating the end of the sentence and leading to the terminate state. The weights of the LSTM are the policy parameters denoted by $\theta$.  More specifically, given the current state $s_t = S_{1:t} = \{x_1, x_2, \cdots, x_t\}$, where $x_i \in \mathcal{V}$ is the token generated at step $t$, the LSTM will output a sequence of hidden states $\{h_1, h_2, \cdots, h_t\}$ by the following recursive functions:
 \begin{align}
     h_t = {\rm LSTM}(x_t, h_{t-1})
 \end{align}
With the hidden states $\{h_1, h_2, \cdots, h_t\}$, the representation of current state $m_t$ can be obtained through:
 \begin{align}
    m_t = g(h_1, h_2, \cdots, h_t)
 \end{align}
\noindent where $g(\cdot)$ is a function that maps the hidden states to a $d$-dimensional hidden space and there are many choices for $g(\cdot)$ such as using a popular attention mechanism~\cite{bahdanau2014neural}. In this work, we let $g(h_1, h_2, \cdots, h_t)=h_t$, which is commonly used for many RNN-based models and leave the investigation into other choices as a future work. With the representation $m_t$, the probability distribution over all the possible actions $\pi_{\theta}(a|m_t)$ is calculated as:
 \begin{align}
     \pi_{\theta}(a|m_t) = {\rm softmax}(W_v m_t + b_v)
 \end{align}
\noindent where $W_v \in \mathbb{R}^{|\mathcal{V}| \times d}$ and $b_v$ are the parameters to be learned during the training process when having a vocabulary size of $|\mathcal{V}|$. 
 
\subsubsection{Reward}
Remember that $S^T$ is the target sentence and let $S^D = D(S)$ be the response produced by the dialogue model $D$ when given $S$ as input. We define two different reward functions for the two manipulation tasks.

\textbf{Target word task.} In this task we want to learn an input sentence which leads to the dialogue model to output a response containing the target word $w^T$. Thus, we wish that the probability of the target word predicted by the output layer of the dialogue model to be the largest at some timestamp $t \in [T]$. We define the reward of the input sentence $S$ as follows:
\begin{align}
    \label{eq:1_reward}
    R(S) = \max_{t \in [T]} (p_t(w^T) - \max_{w \neq w^T}\{p_t(w^T)\})
\end{align}
\noindent where $p_t(w^T)$ is the predicted probability of target word $w^T$ at timestamp $t$ and $\max_{w \neq w^T}\{p_t(w^T)\}$ indicates the largest probability among all the words other than $w^T$ at this timestamp.


\textbf{Target response task.} In this task, we wish the dialogue model can output a response that is semantically similar to the target response but not necessarily exactly matching it. We use the average word embedding similarity \cite{DBLP:journals/corr/WietingBGL15a} between $S^T$ and $S^D=D(S)$ as the reward for the input sentence generated by the policy network. 
We define the reward as follows:
 \begin{align}
    \label{eq:2_reward}
     R(S) = {\rm Sim}(S^T, D(S))
 \end{align}
\noindent where $S$ is the sentence generated by the policy network and $Sim$ is the similarity measurement.

Eq~(\ref{eq:1_reward}) and (\ref{eq:2_reward}) only provide the final rewards to the whole generated sentence $S$. In addition, we design the reward $R(S_{1:t}, a_t)$ at the intermediate state $s_t$ to be the action-value function. Formally, 
\begin{align}
    R(S_{1:t},a_t) = Q^{\pi_\theta}(S_{1:t}, a_t)
\end{align}
\noindent where $Q^{\pi_\theta}(a, S_t)$ is the Q-value, which estimates the expected final reward for the given action and state pair~\cite{watkins1992q}. We further adopt the Monte Carlo methods~\cite{yu2017seqgan,sutton2000policy} to estimate $Q^{\pi_\theta}(S_{1:t},a_t)$.
The way to calculate the intermediate reward is shown in Figure \ref{fig:rewards}.
Specifically, given $S_{1:t}$ and $a_t$, we use current policy $\pi_\theta$ to continuously simulate $T-t$ tokens until the terminate state is reached, where $T$ is the length of the complete simulated sentence. The simulation process will be repeated $N^r$ times. We define this formally as follows:
\begin{align}
    \left \{  S^j_{1:T}\right \}_{j=1}^{N^r} = {\rm Simulation}^{\pi_\theta}(S_{1:t}, a_t)
\end{align}
\noindent where $S^j_{1:T}$ is a simulated sentence and $S^j_{1:t} = S_{1:t}$ for all $j \in \{1, \dots ,N^r \}$. Now, given $\left \{  S^j_{1:T}\right \}_{j=1}^{N^r}$, the estimation of $R(S_{1:t},a_t)$ is calculated as:
\begin{align}
    \label{eq:s_reward}
    R(S_{1:t},a_t) = \frac{1}{N^r} \sum_{j=1}^{N^r}  {\rm Sim}(S^T, D(S^j_{1:T}))
\end{align}
\noindent where $t < T$ indicates the intermediate step. We note that when $t = T$ we can directly use Eq~(\ref{eq:1_reward}) and (\ref{eq:2_reward}) rather than Eq.~(\ref{eq:s_reward}).

\noindent{\bf Objective Function.} 
With the previously defined reward function, the objective function that the agent aims to maximize can be formulated as follows:
\begin{align}
    \label{eq:obj_1}
     J(\theta) = \mathbb{E}_{S\sim \pi_\theta(S)} R(S) 
\end{align}
The accurate value of $J(\theta)$ in Eq.~(\ref{eq:obj_1}) is very difficult to obtain in practice. Next, we describe our optimization procedure for TDGPN.

\subsection{Monte-Carlo Policy Gradient: REINFORCE}
To optimize the objective in Eq.~(\ref{eq:obj_1}), we adopt the widely used REINFORCE algorithm~\cite{williams1992simple}, where Monte-Carlo sampling is applied to estimate $\nabla_\theta J(\theta)$. Specifically, 
\begin{align}
    \label{eq:gradient_1}
    \nabla_\theta J(\theta) &= \sum_{S} R(S) \nabla \pi_\theta(S) \\ \nonumber
    & = \mathbb{E}_{S\sim\pi_\theta(S)}[R(S) \pi_\theta(S) \nabla \log\pi_\theta (S)] \\\nonumber
    & \approx \frac{1}{N}\sum_{i=1}^N \sum_{t=1}^T \nabla_\theta \log \pi_\theta (a^i_t|S^i_{1:t}) R(S)\nonumber
\end{align}
We replace the reward for the whole sentence in Eq.~(\ref{eq:gradient_1}) with the intermediate reward to accelerate the training speed and effectiveness~\cite{liu2017improved}. Thus, the estimated gradient of TDGPN's objective is rewritten as:
\begin{align}
    \label{eq:gradient_2}
    \nabla_\theta J(\theta) &\approx \frac{1}{N}\sum_{i=1}^N \sum_{t=1}^T \nabla_\theta \log \pi_\theta (a^i_t|S^i_{1:t})R(S^i_{1:t}, a_t^i)
\end{align}
With the obtained gradient $\nabla_\theta J(\theta)$, the parameters of the policy network $\pi_\theta$ can be updated as follows:
\begin{align}
    \label{eq:update}
    \theta = \theta + \alpha  \nabla_\theta J(\theta)
\end{align}
where $\alpha$ is the learning rate. 

\subsection{Alternate Learning}

At the beginning of the RL training, since the policy network has little knowledge about the environment, it just randomly generates input sentences which hardly lead the dialogue model to output an expected response. As a result, few action-state pairs $(S_{1:t, a_t})$ can receive positive rewards $R(S_{1:t, a_t})$. This problem impacts training efficiency seriously. To solve this issue, we propose to use samples that we know are closer to the desired input sentences so far as extra guidance to train the policy network. We call them ``pacesetter'' samples. Recall that in each iteration, we sample $N$ input sentences for RL training. The lengths of these sample sentences are 
$\{t_i\}_{i=1}^{N}$. To estimate the reward for each state-action pair, we simulate $N^r$ complete input sentences. So we can get totally $\sum_{i=1}^N N^r t_i$ complete input sentences and their corresponding rewards. We collect the input sentences with the largest top-$K$ rewards as the pacesetter samples.  Then, we use the pacesetter samples as the training examples to optimize the policy network via supervised learning (SL) for once. We update the parameters in the policy network by optimizing the maximum likelihood estimation (MLE) objective of the pacesetter samples. In this way, we perform RL and SL alternately to speed up the convergence of the policy network. 

The detailed optimization process is shown in Algorithm~\ref{alg:reinforce}.
Here we briefly introduce the algorithm.
In line 1, we initialize the policy network $\pi_\theta$ using a pre-trained language model $\pi_{\theta_0}$.
From line 3 to line 10, we use our proposed algorithm to update the parameters of $\pi_\theta$, until it could generate sentences that will make the dialogue model output the target sentence $S^T$ or the number of iterations exceed $M$.
Specifically, in line 5, we randomly sample $N$ sentences from $\pi_\theta$, then for each sampled sentence at each step $t$, we sample $N^r$ sentences to estimate the intermediate reward (i.e., action-value) $R(S^i_{1:t}, a^i_t)$ and compute the gradient according to Eq.~(\ref{eq:gradient_2}),  which is used to update $\theta$ by Eq.~(\ref{eq:update}). 
In lines 14 and 15, we collect the pacesetter samples and update the policy network on them via supervised learning.
If the policy network cannot find a desired sentence $S^*$ within $M$ iterations, the algorithm will output a ``Failure''.

\SetAlCapNameFnt{\small}
\SetAlCapFnt{\small}
\DontPrintSemicolon
\begin{algorithm}[h]\small

\KwIn{Dialogue model $D$, a pre-trained language model $\pi_{\theta_0}$, target word $w^T$ or target response $S^T$, hyper-parameters including $N$, $M$, $N^r$, $K$ and $\alpha$} 
\KwOut{an desired sentence $S^*$ or ``Failure''}
Initialize $\pi_\theta$: $\pi_\theta \gets \pi_{\theta_0}$;\\
iter $\gets$ 1;

\Repeat{iter $>=M$ or find a sentence $S^*$ such that $D(S^*)$ satisfies the requirement.}
    {
    iter $\gets$ iter + 1;\\
    Generate $N$ sentences $\left \{ S^i\right\}_{i=1}^{N}$ from $\pi_\theta$\;
    \For{$i\gets1$ \KwTo $N$ }{
        \For{$t\gets1$ \KwTo $T$}{ Sample {$\left\{S^{ij}_{t}\right\}_{j=1}^{N^r}$} based on $S^{i}_{1:t}$ and $\pi_\theta$\;
        Compute $R(S^i_{1:t}, a^i_t)$ according to Eq.~(\ref{eq:s_reward})\;
        
        $\nabla_\theta J(\theta) \leftarrow  \nabla_\theta J(\theta) + \nabla_\theta log \pi_\theta(a^i_t, S^i_{1:t})R(S^i_{1:t}, a_t^i)$\;
        }
    }
    $\theta \leftarrow \theta +  \frac{\alpha}{N}\nabla_\theta J(\theta)$;\\
    Collect the top-$K$ pacesetter samples $\{p_i\}_{i=1}^{K}$ from $\{S_t^{ij}|i \in [1, N], j \in [1, N^r], t \in [1, t_i]\}$;\\
    Update $\pi_\theta$ on $\{p_i\}_{i=1}^{K}$;
    
  }
    
\caption{{\bf Optimization for TDGPN} \label{alg:reinforce}}
\end{algorithm}
\section{Experiment}
\label{sec:exp}

\subsection{Experimental Settings}
In this subsection, we give a description of the experimental settings including the state-of-the-art dialogue model we are trying to manipulate, the implementation details of the policy network, and the training procedure of the proposed framework.

{\bf The Dialogue Model.} The experiments are conducted on a classic Seq2Seq \cite{sutskever2014sequence} neural dialogue model. 
In this Seq2Seq model, both the encoder and the decoder are implemented by 3-layer LSTM networks with hidden states of size 1024. The vocabulary size is 30,000. Pre-trained Glove word vectors~\cite{pennington2014glove} are used as the word embeddings whose dimension is 300. 

{\bf Dataset.} \label{sec:data} A \emph{Twitter Conversation Corpus} is used to pre-train the policy network, construct target response list, etc. This corpus is different from the one for training the black-box dialogue model. The corpus consists of tweets and replies extracted from Twitter. It contains 1,394,939 single-turn English dialogues in open domains. The dataset is randomly split into training set, validation set, and test set, which consist of 976,457, 139,498 and 278,984 dialogues, respectively.

{\bf Implementation Details.} In this work, we use a 2-layer LSTM with the hidden size being 1,024 as the policy network, which is implemented with Pytorch~\cite{paszke2019pytorch}.
Before the manipulation experiments, we pre-trained the LSTM on the training set of the Twitter dialogue corpus where every single post or reply is treated as a sentence, resulting in around 1.9 million sentences in total. 
Specifically, in the pre-training process, the model is optimized by the standard stochastic gradient descent algorithm with the learning rate of 1.0.
In addition, to prevent overfitting issues, we apply the dropout with the rate of 0.1 and gradient clipping with clip-value being 0.25.
Moreover, the word embeddings are randomly initialized and fine-tuned during the pre-training process. 


As for the details in relation to Algorithm~\ref{alg:reinforce}, we set the sample size $N$ to be 5. The number of simulations per generated sentence (i.e., $N^r$) is set to 5. The maximum sequence length is 5 and 10 for target word task and target response task, respectively. In addition, the policy network parameters are optimized by Adam~\cite{kingma2014adam} with the learning rate of $0.001$ and $0.05$ for the above two tasks, respectively. During RL training, once we find a sample that satisfies a requirement, the training process will stop and we consider it to be successful. On the other hand, if the model cannot find a desired sample within $M=50$ iterations, then we consider it as a failure. For the target word task, the requirement is that the output response contains the target word; while for the target response task, the requirement is that the similarity between the output response and the target response exceeds a threshold.

We note in the target word task, we don't directly adopt the reward defined in Eq. (\ref{eq:1_reward}). Instead, for an input sentence $S$, we use $\max(R(S)-b,0)$, where $b$ is a baseline defined as the average reward value of a batch of input sentences. By the function $\max(\cdot,0)$, we replace all the negative rewards with $0$ and only keep the positive rewards. Additionally, for the target response task, before RL training, we build a reference corpus by feeding 200k utterances (from the training set of the Twitter dialogue corpus) into the dialogue model and obtain 200k input-response pairs. Then, in order to improve the efficiency of training, at the beginning of each training session, we search 5 inputs whose responses are most similar to the target response as the first batch to train the model. 





\subsection{Experimental Results}

In this subsection, we present the results of our experiments.

\textbf{Target word task.} In the experiments, we construct two lists of words as our target words. We randomly sample 200 words from the most frequent 800 words of the Twitter conversation corpus. They form the \textbf{Common} target word list. Moreover, we manually construct a \textbf{Malicious} target word list containing 50 malicious or dirty words that a dialogue agent has to avoid to say~\footnote{When doing experiments on the malicious target words, we set $N=10$ and $M=100$.}.

Table \ref{tab:word} shows the success rate of our proposed TDGPN as well as the average number of iterations the algorithm performs to achieve a successful manipulation. We can see that our algorithm can manipulate the dialogue model successfully for a considerable number of target words. Moreover, we observe that the common words achieve higher success rate and need fewer iterations, which means that it's easier to manipulate the common words than the malicious words. We show three cases of successful manipulation on malicious target words in Table \ref{tab:word_case}. 



\begin{table}
  \caption{Results of the target word task.}
  \label{tab:word}
  \centering
  \begin{tabular}{c|cc}
    \toprule
     & Success Rate & Average Iterations\\
    \midrule
    Common & 65\% & 12.64 \\
    \hline
    Malicious & 30\% & 38.73 \\
    \bottomrule
  \end{tabular}
\end{table}

\begin{table}[!t]
\centering
  \small
  \caption{Case Study of the target word task on Malicious target word list. Input indicates the input crafted by TDGPN and output is the corresponding response produced by the dialogue model. Num.Iter represents the number of iterations.}
  \begin{tabular}{l}
  \hline
  \textbf{Target word:} shit\\
  \textbf{Input:} then start to eat\\
  \textbf{Output:} i ' m not going to eat that \textbf{shit}\\
  \textbf{Num.Iter:} 5\\
  \hline
  \textbf{Target word:} ass\\
  \textbf{Input:} fat , i'm too classy\\
  \textbf{Output:} i ' m not a fat \textbf{ass}\\ 
  \textbf{Num.Iter:} 7\\
  \hline
  \textbf{Target word:} idiot\\
  \textbf{Input:} when he is boring that\\
  \textbf{Output:} he ' s a fucking \textbf{idiot}\\ 
  \textbf{Num.Iter:} 24\\
  \hline
  \end{tabular}
  \label{tab:word_case}
    \vskip -2ex
\end{table}

    

\textbf{Target response task.} For the target response task, we first construct two lists of target responses. 
So called \textbf{Generated} and \textbf{Real} target response lists are used.
To construct the generated target response list, we first feed 200k human utterances from the test set of the Twitter dialogue corpus into the dialogue model to get 200k generated responses and then randomly sample 200 responses as the targets in length 1-3, 4-6 and 7-10, respectively.
The real target response list is obtained by randomly sampling sentences from the rest part of the test set of the Twitter dialogue corpus.
And we filter out some sentences which also occur in the generated target response list, to ensure that there is no overlap between two lists.
The number of real target responses in each length group is also 200.

\begin{figure}[t]
\begin{center}

\hspace*{-0.65cm}  
\includegraphics[scale=0.7]{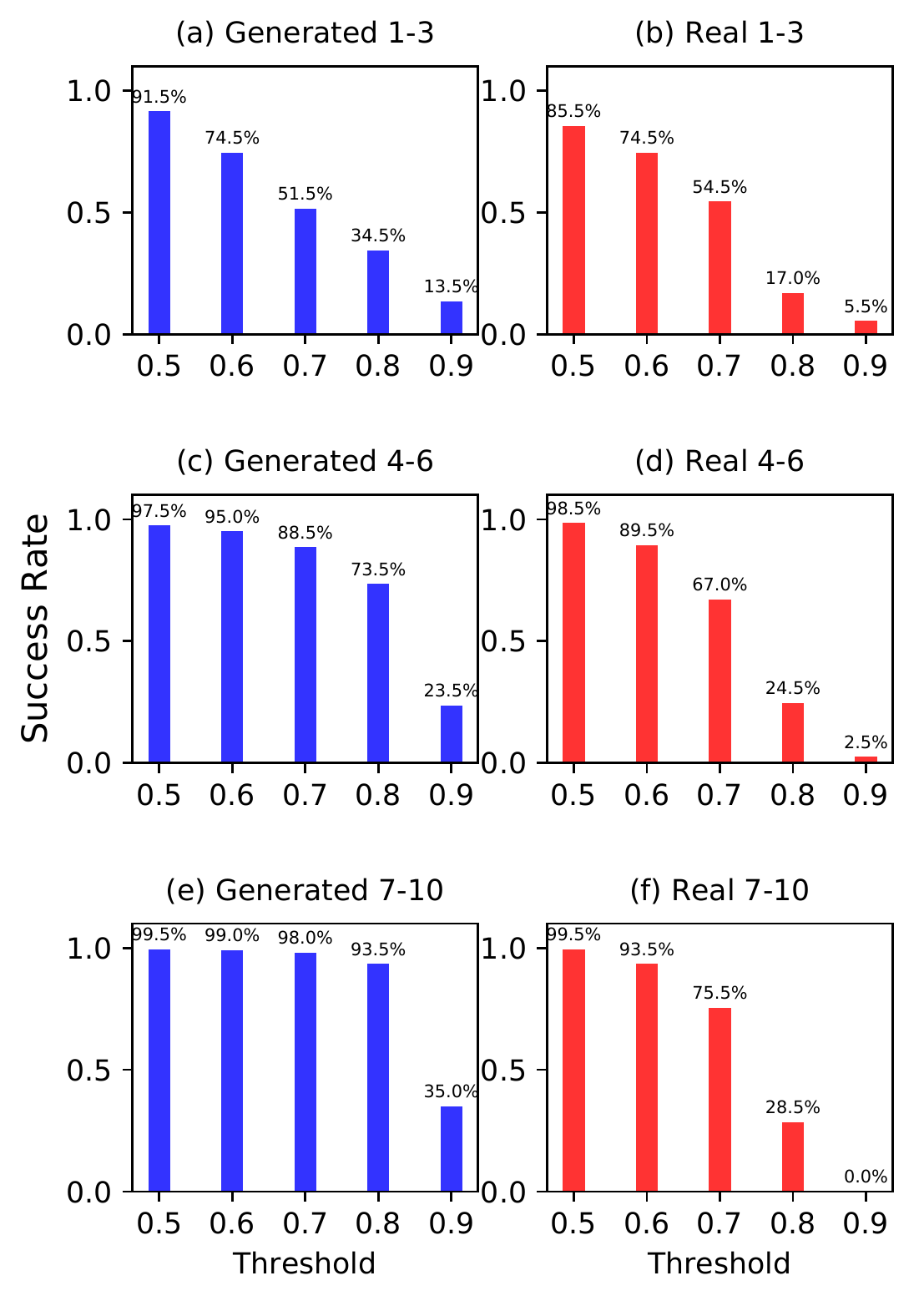}
\end{center}

\vspace{-2ex}
\caption{Results for the target response task.} 
\label{fig:results}
\vskip -3ex
\end{figure}

In Figure~\ref{fig:results}, we show the success rates of TDGPN for manipulating the Twitter dialogue model on two lists of target responses.
The figure shows how success rates vary with different thresholds.
For example, in Figure (a), we can successfully find desired inputs that lead to responses whose similarities with the target ones are above the threshold $0.8$, for $34.5\%$ generated target responses with length 1-3. 


First of all, from the figures we can see that for both the generated and the real target lists, TDGPN can achieve a success rate $85\%-100\%$ with a threshold of $0.5$.
Especially for more than around $80\%$ generated targets with length greater than or equal to $4$, TDGPN is able to find desired inputs that lead to a similarity score above $0.8$.
One key observation is that the model performs significantly better on the generated target list than on the real target response list.
Actually, the neural dialogue models suffer from a safe response problem \cite{li2015diversity}.
Such models tend to offer generic responses to diverse inputs, which makes it hard for the model to provide a target specific response (often seen in real human conversations).
In addition, we observe that the success rate of manipulation is closely related to the length of target responses.
Except for a few cases, for longer target responses, it's easier for TDGPN to manipulate the dialogue model to say something similar to them.

{\it The Quality of Crafted Inputs.} For each real target response, TDGPN tries to find an input whose corresponding output response is most similar to the target one. We also feed the {\it real inputs} corresponding to the target responses in the corpus into the dialogue model. We aim to check how similar the output responses to the target ones in manipulated and original settings. To further demonstrate the effectiveness of the proposed framework in manipulating the dialogue model,  we calculate these similarity scores for each real target response and report the average value in Table~\ref{tab:results}.
From the table, we make the following observations.
First, even inputting the real inputs, the similarity scores between the output responses and the target responses are not high.
Second, with the generated inputs from the proposed framework, the similarity scores are significantly improved.
Specifically, the word embedding similarity increases by $57.2\%$, $40.2\%$ and $32.2\%$ for the target responses with length 1-3, 4-6 and 7-10, respectively.  

{\it Case Study.} Table~\ref{tab:case} shows five examples in the manipulating experiments.
The first three target responses are from the generated target response list while the other two are from the real response list.
Given those target responses, desired inputs are successfully crafted.
Each of them leads to an output of the dialogue model similar to the target one, evaluating by the corresponding similarity measurement.
Besides, unlike some related works \cite{he2018detecting,cheng2018seq2sick} where crafted text inputs are ungrammatical and meaningless, the inputs generated by our model are smooth and natural utterances, which is guaranteed by the language model.


\begin{table}
  \caption{Average similarity scores between the output response and the target response in \textbf{Real} list.}
  \label{tab:results}
  \centering
  \begin{tabular}{c|ccc}
    \toprule
    \cmidrule(r){1-2}
    \textbf{Length}     & 1-3     & 4-6 & 7-10 \\
    \midrule
    Real Input & 0.439  & 0.518 & 0.566     \\
    \hline
    TDGPN & 0.69  &  0.726 & 0.748     \\
    \bottomrule
  \end{tabular}
\end{table}

\begin{table*}[!t]
\centering
  \caption{Case Study. The first column shows the inputs from TDGPN. The middle column shows the target responses and the outputs by the dialogue model. The third column shows the similarity score.}
  \begin{tabular}{l|l|l}
  \hline
    \textbf{Inputs} & \textbf{Responses} & \textbf{Similarity}\\
     \hline
     \multirow{2}{5.5cm}{the band perry is goin to be at the movies} &  \textbf{Target}: i ' m going to be there tomorrow ! & 0.958\\
     & \textbf{Output}:  i ' m going to be there & \\
     \hline
     \multirow{2}{5.5cm}{i followed you so you better follow me back} &  \textbf{Target}: i ' m not following you . & 0.952\\
     & \textbf{Output}: i ' m sorry i ' m not following you &  \\
     \hline
     \multirow{2}{5.5cm}{so i have a friend in the sea} &  \textbf{Target}: i ' m in the same boat lol & 0.958\\
     & \textbf{Output}:  i ' m in the same boat . & \\
     \hline
     \multirow{2}{5.5cm}{what's poppin peeps ?} &  \textbf{Target}: nothing much just us chatting shit really & 0.826\\
     & \textbf{Output}: nothing much , just chillin & \\
     \hline
     \multirow{2}{5.5cm}{honestly i miss my brother} &  \textbf{Target}: me = miss you ( : lol . & 0.876\\
     & \textbf{Output}: i miss you too & \\
     \hline
    
  \hline
  \end{tabular}
  \label{tab:case}
    \vskip -2ex
\end{table*}




\section{Related Work}
\label{sec:rela}

Model attack in NLP has been a fast-moving field especially across many neural based methodologies \cite{chakraborty2018adversarial,DBLP:journals/corr/abs-1909-08072}, since our understanding of their inner workings is quite limited. Chen et al.~\cite{chen2018attacking} investigate the ability for small perturbations that could result in the image captioning to contain a randomly selected keyword/caption. Although this is similar to our problem in that they are generating target text, the key difference is that they are working with continuous inputs (i.e., images). Similarly, some work has focused on text classification~\cite{kuleshov2018classification,liang2018fooled}, but in the white-box setting as compared to our framework is proposed in the black-box setting. 

Our work is primarily related to other works focused on building a better understanding of sequence-to-sequence based dialog system models, such as their over-sensitivity and over-stability~\cite{niu2018adversarial}, robustness~\cite{cheng2018seq2sick} and likeliness of generating egregious responses~\cite{he2018detecting}. In~\cite{cheng2018seq2sick} the problem of attacking sequence-to-sequence models is presented to evaluate the robustness of this class of deep neural networks. Unlike our work, they focus on the development of a white-box framework that is built around having the gradient. However, in practice, the assumption of accessing the full knowledge of the neural network is far less practical~\cite{zhang2019survey}. Then Niu et al.~\cite{niu2018adversarial} focus on using adversarial training to investigate both over-sensitivity and over-stability of dialogue models, where the small changes to the input should or should not change the output of the dialogue system, respectively. 
Besides, He et al.~\cite{he2018detecting} focus on learning an input to generate a specific offensive/malicious output of a pre-trained dialogue system. However, our proposed framework is based on the black-box setting (unlike their model, which is under the white-box assumption) which raises significantly higher levels of complexity to develop an optimization process. The work~\cite{liu2019say} investigates the target response task on the black-box setting. The authors train a reusable reverse dialogue generator by reinforcement learning and use it to discover the inputs leading to target outputs through multiple trials.  
\section{Conclusion}
Currently, the state-of-the-art dialogue systems are harnessing the power of deep neural models, and although they are proving to become more and more human-like, recent concerns have been raised for neural models across all domains as to whether they can be manipulated (with most focusing on malicious attacks). Thus, in this work, we investigate whether current neural dialogue models can be manipulated and develop a reinforcement learning based sentence generation framework that can learn to craft the input sentences causing dialogue models to output target words and responses. We conduct extensive experiments on a state-of-the-art dialogue neural model and the results show that dialogue systems can indeed be manipulated. In addition, our proposed method is not only able to manipulate neural dialogue model, but it's also likely to be applied on black-box dialogue systems based on other methods (e.g. rule-based, retrieval-based, etc.), or even models for other natural language generation tasks (e.g. text summarization, machine translation, etc.).
We will leave the investigations on these areas as future works.

\bibliographystyle{acl_natbib}
\bibliography{main}

\begin{thebibliography}{42}
\expandafter\ifx\csname natexlab\endcsname\relax\def\natexlab#1{#1}\fi

\bibitem[{Bahdanau et~al.(2014)Bahdanau, Cho, and Bengio}]{bahdanau2014neural}
Dzmitry Bahdanau, Kyunghyun Cho, and Yoshua Bengio. 2014.
\newblock Neural machine translation by jointly learning to align and
  translate.
\newblock \emph{arXiv preprint arXiv:1409.0473}.

\bibitem[{Chakraborty et~al.(2018)Chakraborty, Alam, Dey, Chattopadhyay, and
  Mukhopadhyay}]{chakraborty2018adversarial}
Anirban Chakraborty, Manaar Alam, Vishal Dey, Anupam Chattopadhyay, and Debdeep
  Mukhopadhyay. 2018.
\newblock Adversarial attacks and defences: A survey.
\newblock \emph{arXiv preprint arXiv:1810.00069}.

\bibitem[{Chen(2018)}]{chen2018attacking}
Hongge Chen. 2018.
\newblock Attacking visual language grounding with adversarial examples: A case
  study on neural image captioning.
\newblock In \emph{ACL}.

\bibitem[{Chen et~al.(2017)Chen, Liu, Yin, and Tang}]{chen2017survey}
Hongshen Chen, Xiaorui Liu, Dawei Yin, and Jiliang Tang. 2017.
\newblock A survey on dialogue systems: Recent advances and new frontiers.
\newblock \emph{ACM SIGKDD Explorations Newsletter}, 19(2):25--35.

\bibitem[{Cheng et~al.(2018)Cheng, Yi, Zhang, Chen, and
  Hsieh}]{cheng2018seq2sick}
Minhao Cheng, Jinfeng Yi, Huan Zhang, Pin-Yu Chen, and Cho-Jui Hsieh. 2018.
\newblock Seq2sick: Evaluating the robustness of sequence-to-sequence models
  with adversarial examples.
\newblock \emph{arXiv preprint arXiv:1803.01128}.

\bibitem[{Dinan et~al.(2019)Dinan, Fan, Williams, Urbanek, Kiela, and
  Weston}]{dinan2019queens}
Emily Dinan, Angela Fan, Adina Williams, Jack Urbanek, Douwe Kiela, and Jason
  Weston. 2019.
\newblock Queens are powerful too: Mitigating gender bias in dialogue
  generation.
\newblock \emph{arXiv preprint arXiv:1911.03842}.

\bibitem[{Goddeau et~al.(1996)Goddeau, Meng, Polifroni, Seneff, and
  Busayapongchai}]{goddeau1996form}
David Goddeau, Helen Meng, Joseph Polifroni, Stephanie Seneff, and Senis
  Busayapongchai. 1996.
\newblock A form-based dialogue manager for spoken language applications.
\newblock In \emph{Proceeding of Fourth International Conference on Spoken
  Language Processing. ICSLP'96}, volume~2, pages 701--704. IEEE.

\bibitem[{Goodfellow et~al.(2014)Goodfellow, Shlens, and
  Szegedy}]{goodfellow2014explaining}
Ian~J Goodfellow, Jonathon Shlens, and Christian Szegedy. 2014.
\newblock Explaining and harnessing adversarial examples.
\newblock \emph{arXiv preprint arXiv:1412.6572}.

\bibitem[{He and Glass(2018)}]{he2018detecting}
Tianxing He and James Glass. 2018.
\newblock Detecting egregious responses in neural sequence-to-sequence models.
\newblock \emph{arXiv preprint arXiv:1809.04113}.

\bibitem[{Henderson et~al.(2017)Henderson, Sinha, Angelard-Gontier, Ke, Fried,
  Lowe, and Pineau}]{henderson2017ethical}
Peter Henderson, Koustuv Sinha, Nicolas Angelard-Gontier, Nan~Rosemary Ke,
  Genevieve Fried, Ryan Lowe, and Joelle Pineau. 2017.
\newblock Ethical challenges in data-driven dialogue systems.
\newblock \emph{arXiv preprint arXiv:1711.09050}.

\bibitem[{Hu et~al.(2014)Hu, Lu, Li, and Chen}]{hu2014convolutional}
Baotian Hu, Zhengdong Lu, Hang Li, and Qingcai Chen. 2014.
\newblock Convolutional neural network architectures for matching natural
  language sentences.
\newblock In \emph{Advances in neural information processing systems}, pages
  2042--2050.

\bibitem[{Kingma and Ba(2014)}]{kingma2014adam}
Diederik~P Kingma and Jimmy Ba. 2014.
\newblock Adam: A method for stochastic optimization.
\newblock \emph{arXiv preprint arXiv:1412.6980}.

\bibitem[{Kuleshov et~al.(2018)Kuleshov, Thakoor, Lau, and
  Ermon}]{kuleshov2018classification}
Volodymyr Kuleshov, Shantanu Thakoor, Tingfung Lau, and Stefano Ermon. 2018.
\newblock Adversarial examples for natural language classification problems.
\newblock \emph{Open Review submission OpenReview:r1QZ3zbAZ}.

\bibitem[{Li et~al.(2015)Li, Galley, Brockett, Gao, and
  Dolan}]{li2015diversity}
Jiwei Li, Michel Galley, Chris Brockett, Jianfeng Gao, and Bill Dolan. 2015.
\newblock A diversity-promoting objective function for neural conversation
  models.
\newblock \emph{arXiv preprint arXiv:1510.03055}.

\bibitem[{Liang et~al.(2018)Liang, Li, Su, Bian, Li, and Shi}]{liang2018fooled}
Bin Liang, Hongcheng Li, Miaoqiang Su, Pan Bian, Xirong Li, and Wenchang Shi.
  2018.
\newblock Deep text classification can be fooled.
\newblock In \emph{Proceedings of the Twenty-Seventh International Joint
  Conference on Artificial Intelligence, {IJCAI-18}}, pages 4208--4215.
  International Joint Conferences on Artificial Intelligence Organization.

\bibitem[{Liu et~al.(2019{\natexlab{a}})Liu, Dacon, Fan, Liu, Liu, and
  Tang}]{liu2019does}
Haochen Liu, Jamell Dacon, Wenqi Fan, Hui Liu, Zitao Liu, and Jiliang Tang.
  2019{\natexlab{a}}.
\newblock Does gender matter? towards fairness in dialogue systems.
\newblock \emph{arXiv preprint arXiv:1910.10486}.

\bibitem[{Liu et~al.(2019{\natexlab{b}})Liu, Derr, Liu, and Tang}]{liu2019say}
Haochen Liu, Tyler Derr, Zitao Liu, and Jiliang Tang. 2019{\natexlab{b}}.
\newblock Say what i want: Towards the dark side of neural dialogue models.
\newblock \emph{arXiv preprint arXiv:1909.06044}.

\bibitem[{Liu et~al.(2017)Liu, Zhu, Ye, Guadarrama, and
  Murphy}]{liu2017improved}
Siqi Liu, Zhenhai Zhu, Ning Ye, Sergio Guadarrama, and Kevin Murphy. 2017.
\newblock Improved image captioning via policy gradient optimization of spider.
\newblock In \emph{Proceedings of the IEEE international conference on computer
  vision}, pages 873--881.

\bibitem[{Lu and Li(2013)}]{lu2013deep}
Zhengdong Lu and Hang Li. 2013.
\newblock A deep architecture for matching short texts.
\newblock In \emph{Advances in Neural Information Processing Systems}, pages
  1367--1375.

\bibitem[{Niu and Bansal(2018)}]{niu2018adversarial}
Tong Niu and Mohit Bansal. 2018.
\newblock Adversarial over-sensitivity and over-stability strategies for
  dialogue models.
\newblock In \emph{The SIGNLL Conference on Computational Natural Language
  Learning (CoNLL)}.

\bibitem[{Paszke et~al.(2019)Paszke, Gross, Massa, Lerer, Bradbury, Chanan,
  Killeen, Lin, Gimelshein, Antiga et~al.}]{paszke2019pytorch}
Adam Paszke, Sam Gross, Francisco Massa, Adam Lerer, James Bradbury, Gregory
  Chanan, Trevor Killeen, Zeming Lin, Natalia Gimelshein, Luca Antiga, et~al.
  2019.
\newblock Pytorch: An imperative style, high-performance deep learning library.
\newblock In \emph{Advances in Neural Information Processing Systems}, pages
  8024--8035.

\bibitem[{Pennington et~al.(2014)Pennington, Socher, and
  Manning}]{pennington2014glove}
Jeffrey Pennington, Richard Socher, and Christopher Manning. 2014.
\newblock Glove: Global vectors for word representation.
\newblock In \emph{Proceedings of the 2014 conference on empirical methods in
  natural language processing (EMNLP)}, pages 1532--1543.

\bibitem[{Price(2016)}]{price2016microsoft}
Rob Price. 2016.
\newblock Microsoft is deleting its ai chatbot’s incredibly racist tweets.
\newblock \emph{Business Insider}.

\bibitem[{Serban et~al.(2016)Serban, Sordoni, Bengio, Courville, and
  Pineau}]{serban2016building}
Iulian~Vlad Serban, Alessandro Sordoni, Yoshua Bengio, Aaron~C Courville, and
  Joelle Pineau. 2016.
\newblock Building end-to-end dialogue systems using generative hierarchical
  neural network models.
\newblock In \emph{AAAI}, volume~16, pages 3776--3784.

\bibitem[{Shang et~al.(2015)Shang, Lu, and Li}]{shang2015neural}
Lifeng Shang, Zhengdong Lu, and Hang Li. 2015.
\newblock Neural responding machine for short-text conversation.
\newblock \emph{arXiv preprint arXiv:1503.02364}.

\bibitem[{Shum et~al.(2018)Shum, He, and Li}]{shum2018eliza}
Heung-Yeung Shum, Xiao-dong He, and Di~Li. 2018.
\newblock From eliza to xiaoice: challenges and opportunities with social
  chatbots.
\newblock \emph{Frontiers of Information Technology \& Electronic Engineering},
  19(1):10--26.

\bibitem[{Sutskever et~al.(2014)Sutskever, Vinyals, and
  Le}]{sutskever2014sequence}
Ilya Sutskever, Oriol Vinyals, and Quoc~V Le. 2014.
\newblock Sequence to sequence learning with neural networks.
\newblock In \emph{Advances in neural information processing systems}, pages
  3104--3112.

\bibitem[{Sutton et~al.(2000)Sutton, McAllester, Singh, and
  Mansour}]{sutton2000policy}
Richard~S Sutton, David~A McAllester, Satinder~P Singh, and Yishay Mansour.
  2000.
\newblock Policy gradient methods for reinforcement learning with function
  approximation.
\newblock In \emph{Advances in neural information processing systems}, pages
  1057--1063.

\bibitem[{Szegedy et~al.(2013)Szegedy, Zaremba, Sutskever, Bruna, Erhan,
  Goodfellow, and Fergus}]{szegedy2013intriguing}
Christian Szegedy, Wojciech Zaremba, Ilya Sutskever, Joan Bruna, Dumitru Erhan,
  Ian Goodfellow, and Rob Fergus. 2013.
\newblock Intriguing properties of neural networks.
\newblock \emph{arXiv preprint arXiv:1312.6199}.

\bibitem[{Vinyals and Le(2015)}]{vinyals2015neural}
Oriol Vinyals and Quoc Le. 2015.
\newblock A neural conversational model.
\newblock \emph{arXiv preprint arXiv:1506.05869}.

\bibitem[{Wallace(2001)}]{wallace2001artificial}
Richard Wallace. 2001.
\newblock Artificial linguistic internet computer entity (alice).

\bibitem[{Watkins and Dayan(1992)}]{watkins1992q}
Christopher~JCH Watkins and Peter Dayan. 1992.
\newblock Q-learning.
\newblock \emph{Machine learning}, 8(3-4):279--292.

\bibitem[{Weizenbaum(1966)}]{weizenbaum1966eliza}
Joseph Weizenbaum. 1966.
\newblock Eliza—a computer program for the study of natural language
  communication between man and machine.
\newblock \emph{Communications of the ACM}, 9(1):36--45.

\bibitem[{Wieting et~al.(2016)Wieting, Bansal, Gimpel, and
  Livescu}]{DBLP:journals/corr/WietingBGL15a}
John Wieting, Mohit Bansal, Kevin Gimpel, and Karen Livescu. 2016.
\newblock \href {http://arxiv.org/abs/1511.08198} {Towards universal
  paraphrastic sentence embeddings}.
\newblock In \emph{4th International Conference on Learning Representations,
  {ICLR} 2016, San Juan, Puerto Rico, May 2-4, 2016, Conference Track
  Proceedings}.

\bibitem[{Williams(1992)}]{williams1992simple}
Ronald~J Williams. 1992.
\newblock Simple statistical gradient-following algorithms for connectionist
  reinforcement learning.
\newblock \emph{Machine learning}, 8(3-4):229--256.

\bibitem[{Wolf et~al.(2017)Wolf, Miller, and Grodzinsky}]{wolf2017we}
Marty~J Wolf, K~Miller, and Frances~S Grodzinsky. 2017.
\newblock Why we should have seen that coming: Comments on microsoft's tay
  experiment, and wider implications.
\newblock \emph{ACM SIGCAS Computers and Society}, 47(3):54--64.

\bibitem[{Xu et~al.(2019)Xu, Ma, Liu, Deb, Liu, Tang, and
  Jain}]{DBLP:journals/corr/abs-1909-08072}
Han Xu, Yao Ma, Haochen Liu, Debayan Deb, Hui Liu, Jiliang Tang, and Anil~K.
  Jain. 2019.
\newblock \href {http://arxiv.org/abs/1909.08072} {Adversarial attacks and
  defenses in images, graphs and text: {A} review}.
\newblock \emph{CoRR}, abs/1909.08072.

\bibitem[{Yu et~al.(2017)Yu, Zhang, Wang, and Yu}]{yu2017seqgan}
Lantao Yu, Weinan Zhang, Jun Wang, and Yong Yu. 2017.
\newblock Seqgan: Sequence generative adversarial nets with policy gradient.
\newblock In \emph{Thirty-First AAAI Conference on Artificial Intelligence}.

\bibitem[{Yuan et~al.(2019)Yuan, He, Zhu, and Li}]{yuan2019adversarial}
Xiaoyong Yuan, Pan He, Qile Zhu, and Xiaolin Li. 2019.
\newblock Adversarial examples: Attacks and defenses for deep learning.
\newblock \emph{IEEE transactions on neural networks and learning systems}.

\bibitem[{Zhang et~al.(2019)Zhang, Sheng, and Alhazmi}]{zhang2019survey}
Wei~Emma Zhang, Quan~Z Sheng, and Ahoud Abdulrahmn~F Alhazmi. 2019.
\newblock Generating textual adversarial examples for deep learning models: A
  survey.
\newblock \emph{arXiv preprint arXiv:1901.06796}.

\bibitem[{Zhao et~al.(2017)Zhao, Dua, and Singh}]{zhao2017generating}
Zhengli Zhao, Dheeru Dua, and Sameer Singh. 2017.
\newblock Generating natural adversarial examples.
\newblock \emph{arXiv preprint arXiv:1710.11342}.

\bibitem[{Zhou et~al.(2018)Zhou, Gao, Li, and
  Shum}]{DBLP:journals/corr/abs-1812-08989}
Li~Zhou, Jianfeng Gao, Di~Li, and Heung{-}Yeung Shum. 2018.
\newblock \href {http://arxiv.org/abs/1812.08989} {The design and
  implementation of xiaoice, an empathetic social chatbot}.
\newblock \emph{CoRR}, abs/1812.08989.

\end{thebibliography}

\end{document}